\documentclass{article}

\usepackage[final]{corl_2017}
\usepackage{amsmath}
\makeatletter
\g@addto@macro\normalsize{%
  \setlength\abovedisplayskip{4pt}
  \setlength\belowdisplayskip{4pt}
  \setlength\abovedisplayshortskip{4pt}
  \setlength\belowdisplayshortskip{4pt}
}
\usepackage{graphics} % for pdf, bitmapped graphics files
\usepackage{graphicx}
\usepackage{amsmath} % assumes amsmath package installed
\usepackage{amsfonts}
\usepackage{xcolor}
\usepackage{wrapfig}
\usepackage{float}  % for [H] for figure
\usepackage{subcaption}

\newcommand{\R}{\mathbb R}

\DeclareMathOperator*{\diag}{diag}

\title{Deep Kernels for Optimizing Locomotion Controllers}

\author{
  Rika Antonova\thanks{Both of these authors contributed equally\vspace{-5px}}\\
  Robotics, Perception and Learning, CSC \\
  KTH Royal Institute of Technology \\ 
  Stockholm, Sweden\\
  \texttt{antonova@kth.se} \\
  \And
  Akshara Rai\footnotemark[1]\\
  Robotics Institute\\
  School of Computer Science\\
  Carnegie Mellon University, USA\\
  \texttt{arai@andrew.cmu.edu} \\
  \And
  Christopher G. Atkeson\\
  Robotics Institute\\
  School of Computer Science\\
  Carnegie Mellon University, USA\\
  \texttt{cga@cs.cmu.edu}
  \vspace{-20px}
}

\begin{document}
\maketitle

%===============================================================================

\begin{abstract}
	{Sample efficiency is important when optimizing parameters of locomotion controllers, since hardware experiments are time consuming and expensive. Bayesian Optimization, a sample-efficient optimization framework, has recently been widely applied to address this problem, but further improvements in sample efficiency are needed for practical applicability to real-world robots and high-dimensional controllers. To address this, prior work has proposed using domain expertise for constructing custom distance metrics for locomotion. In this work we show how to learn such a distance metric automatically. 
	We use a neural network to learn an informed distance metric from data obtained in high-fidelity simulations. We conduct experiments on two different controllers and robot architectures. First, we demonstrate improvement in sample efficiency when optimizing a 5-dimensional controller on the ATRIAS robot hardware. We then conduct simulation experiments to optimize a 16-dimensional controller for a 7-link robot model and obtain significant improvements even when optimizing in perturbed environments. This demonstrates that our approach is able to enhance sample efficiency for two different controllers, hence is a fitting candidate for further experiments on hardware in the future.}
\end{abstract}

% Two or three meaningful keywords should be added here

%===============================================================================

\section{Introduction}
\label{sec:intro}
Bayesian Optimization (BO) is rapidly becoming a popular approach for optimizing controllers in robotics. It offers sample-efficient, black-box and gradient-free optimization, well suited for many problems in the field. Recently, some success has also been achieved when optimizing controllers directly on hardware~\citep{calandra2016AMAI, marco2017virtual, cully2015robots}. Hence, this sample-efficient optimization framework has the potential to alleviate the need for manual tuning by experts, to a large extent. However, for high-dimensional controllers and challenging cost functions the performance of conventional BO often degrades. Without an informative prior, the number of data points required could be prohibitively expensive for hardware-only optimization. Hence, it seems ideal to exploit simulation to speed up learning, as proposed in~\citep{rai2016sample} and ~\citep{cully2015robots}. These prior approaches, however, need extensive expert domain knowledge to define the problem-specific informed distance metric.

In this work we demonstrate how to construct an informed metric automatically, without relying heavily on domain experts. We propose to learn a distance metric with a neural network, utilizing data obtained from a high-fidelity simulator. This involves first running short simulations of a locomotion controller on a large grid of control parameters and recording the behavior of each set of parameters. The neural network then learns a mapping between input controller parameters and simulation output/behavior. We propose two ways of defining the target to be learned by the network. The first approach is based on the cost function that is to be optimized with BO on hardware, or a perturbed simulator. The second is cost-agnostic: learning to reconstruct a summary of robot trajectories obtained from simulation. This provides a useful re-parameterization: controller parameters that produce similar walking trajectory summaries are closer in this re-parameterized space. 

\begin{wrapfigure}{r}{0.31\textwidth}
\vspace{-7px}
\includegraphics[width=0.3\textwidth]{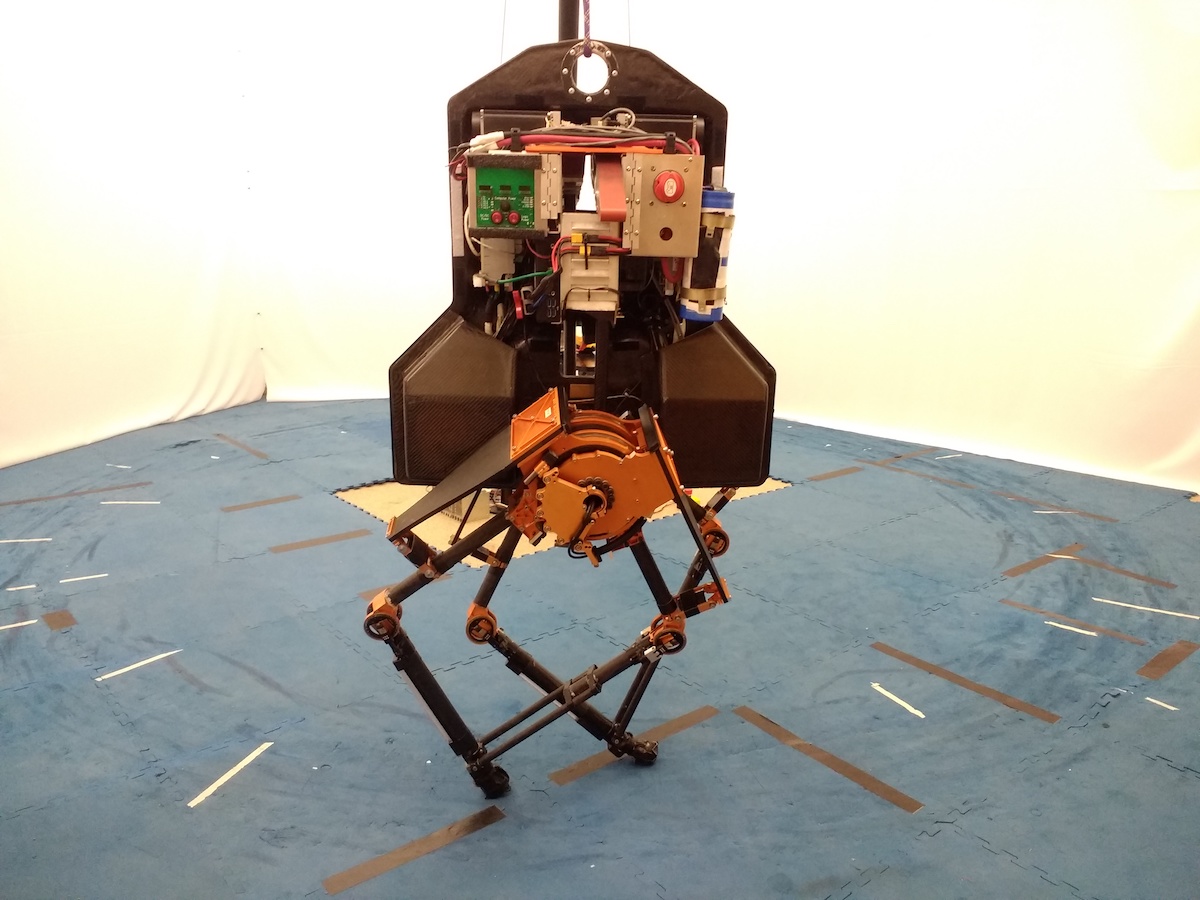}
\includegraphics[width=0.3\textwidth]{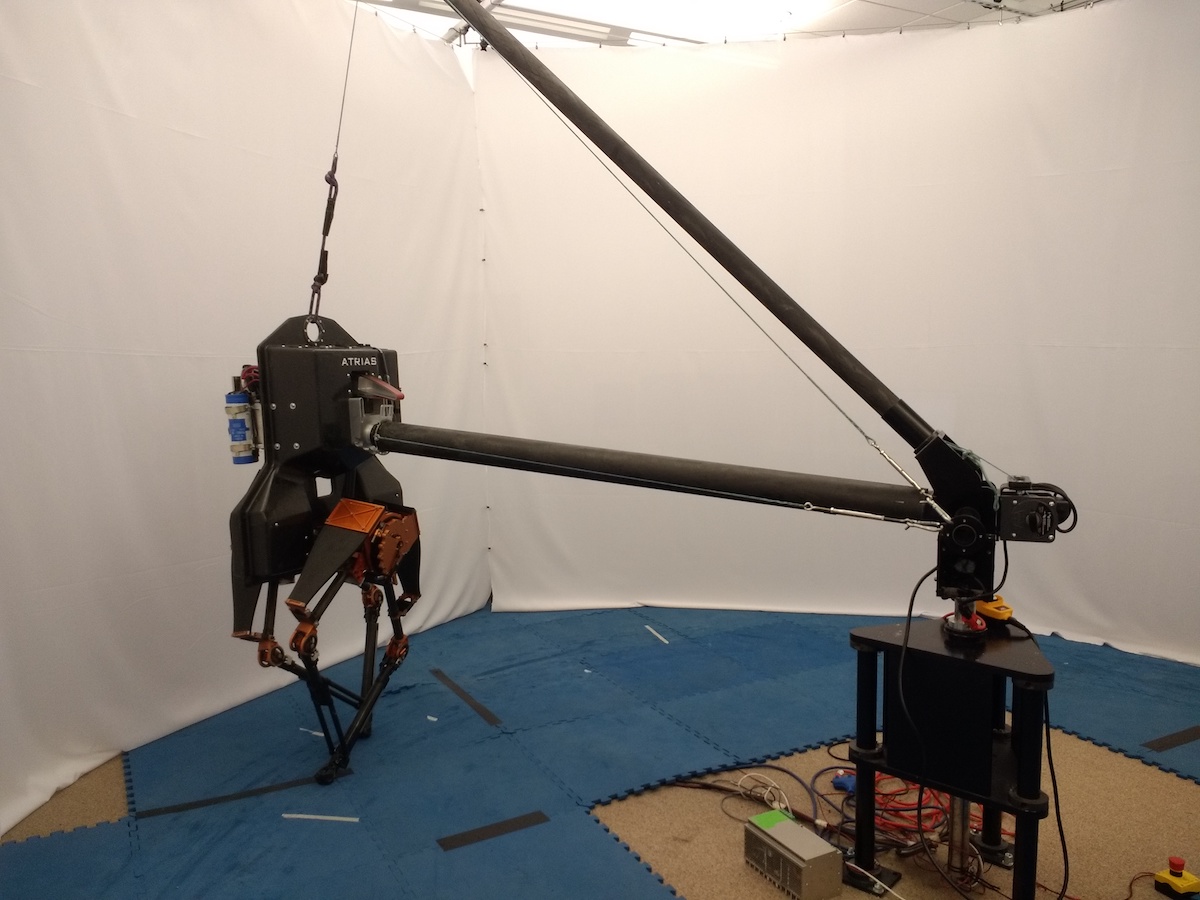}
\vspace{-5px}
\caption{\small{ATRIAS robot. \quad \quad }}
\label{fig:atrias}
\vspace{-13px}
\end{wrapfigure}
In our first set of experiments we optimize a 5-dimensional controller on the ATRIAS robot hardware (\mbox{Figure~\ref{fig:atrias}}). We demonstrate that using  cost-based kernel obtained with our approach outperforms using an uninformed kernel for BO. The setting we consider for ATRIAS experiments yields a 
proof-of-concept rather than a large-scale optimization problem. Nonetheless, we believe that its an important step towards optimizing locomotion policies for complex humanoid robots on hardware.

Prior Bayesian Optimization studies often used simpler robots. For example, \citep{tesch} use snake robots, \citep{cully2015robots} use a hexapod, which often have statically stable gaits, or spend a significant amount of gait duration in a statically stable state. \citep{calandra2016AMAI} use a smaller biped with a finite-state-machine controller, which is not widely used. In contrast, ATRIAS is a complex bipedal robot, which cannot be statically stable in single support because of point feet. Hence, it is likely to fall with unstable controllers. Moreover, our control framework is in line with most state-of-the-art robot controllers \citep{feng2015optimization}, \citep{kuindersma2016optimization}. Hence, results on our testbed can be transferred to other systems.

Our second set of experiments is on the Neuromuscular model~\citep{geyer2010muscle}. We optimize a 16-dimensional controller for a 7-link robot model in simulation. Our approach of reconstructing trajectory summaries again yields a significant improvement over using uninformed kernels for BO.
This is the case for both a smooth and a challenging non-smooth cost suggested in prior literature~\citep{song2015neural}. Hence the proposed approach offers a promising way to construct cost-agnostic kernels for BO automatically.

%===============================================================================

\section{Background}
\label{sec:background}
\subsection{Optimizing Bipedal Locomotion Controllers}

Approaches to optimizing locomotion controllers range from manual tuning to fully automatic optimization. For complex controllers fully manual tuning is sometimes infeasible or excessively time consuming. In such cases, approaches like \mbox{CMA-ES} have been used to find points yielding good performance in simulation first~\citep{song2015neural}. A domain expert could then use such points as starting points to later manually adjust the parameters such that they are effective on hardware. 

Recently there has been significant interest in developing methods for automatic parameter optimization. Bayesian Optimization has been suggested as one of the promising approaches due its sample efficiency. However, it still can take 30-40 samples to optimize a 4 dimensional controller~\citep{calandra2016AMAI}. To enhance kernel flexibility~\cite{calandra2016manifold} suggests supervised learning of a feature transform during regression. However, this approach does not directly support incorporating a very large amount of data from simulation. Even if it is extended to pre-train on simulated data, it is not clear whether further joint optimization would be desirable: 10-100 hardware samples might not be enough to meaningfully affect the transform built from hundreds of thousands of points from simulation.

Recent works proposed using simulation to aid learning on hardware, for example \citep{marco2017virtual}, \citep{cully2015robots}, \citep{rai2016sample}.\\
\citep{marco2017virtual} propose adding noisy evaluations from simulation to BO posterior directly. The limitation is the need to carefully balance the influence of the samples obtained from simulation versus hardware. \\
\citep{cully2015robots} tabulate best performing points versus their average score on a behavioural metric -- average contact time of their hexapod system in simulation. This metric guides trial-and-error learning to quickly find behaviours that compensate for damage of the robot.
The search is done in behaviour space, and limited to pre-selected ``successful'' points from simulation. This helps make their search faster and potentially safer. However, if an optimal point was not pre-selected,  BO cannot sample it during optimization. ``Best points" are cost-specific (the map needs to be re-generated for each cost) and problem specific, so expert-knowledge is needed to apply the method to other systems. \\
\citep{rai2016sample} propose a new distance metric using domain knowledge about bipedal locomotion. Short simulations are used to compute this metric for a large number of points (sets of control parameters), thus distinguishing points based on their behaviour in simulation, rather than the Euclidean distance between them. 
The method generalizes to different costs and locomotion controllers. However, the distance metric is specifically designed for bipedal locomotion. Further domain-specific expertise would be needed to adapt this approach to other settings.

Another recent direction for learning locomotion controllers utilized deep neural networks. ~\citep{peng2016terrain} formulates the problem of learning locomotion gaits as actor-critic Reinforcement Learning with neural networks as function approximators for policy and value functions. However, it is not straightforward to make such approaches data-efficient enough for real hardware  (\citep{peng2016terrain} uses 10 million state-action transitions for training). So in our work we are interested in combining sample efficiency of an approach like Bayesian Optimization with the flexibility and scalability of deep neural networks.

\subsection{Background on Bayesian Optimization}
\label{subsec:bo_background}

Bayesian Optimization is a framework for sample-efficient global search (\citep{BOtutorial2016} gives a recent overview). The goal is to find $\pmb{x}^*$ that optimizes a given objective function $f(\pmb{x})$, while executing as few evaluations of $f$ as possible. In order to select the most promising points to evaluate next, an ``acquisition'' function is defined. One example is Expected Improvement (EI) function that selects $\pmb{x}$ to maximize expected improvement over the value of the best result obtained so far~\citep{mockus1978ei}. EI requires defining the prior/posterior mean and variance of $f$, and Gaussian Process is frequently used for this:
\vspace{-10px}
\begin{equation*}
f(\pmb{x}) \sim \mathcal{GP}(\mu(\pmb{x}), k(\pmb{x}_i, \pmb{x}_j))
\end{equation*}
Here $\mu$ is a mean function and $k$ defines a kernel. $k(\pmb{x}_i, \pmb{x}_j)$ encodes the similarity of two inputs $\pmb{x}_i, \pmb{x}_j$. The value of $f(\pmb{x}_i$) has a significant influence on the posterior value of $f(\pmb{x}_j)$ if $\pmb{x}_i, \pmb{x}_j$ have high similarity according to the kernel. Squared Exponential kernel is widely used:
\begin{equation*}
k_{SE}(\pmb{x}_i, \pmb{x}_j) = \sigma_k^2 \exp\big(- \tfrac{1}{2} (\pmb{x}_i - \pmb{x}_j)^T \diag(\pmb{\ell})^{\!-\!2} (\pmb{x}_i - \pmb{x}_j) \big),
\end{equation*}
where hyperparameters: $\sigma_k^2, \ \pmb{\ell}$ are signal variance and a vector of length scales respectively. It is customary to adjust these automatically during optimization to learn the overall variance and how quickly $f$ varies in each input dimension.

Gaussian Process conditioned on evidence represents a posterior distribution for $f$. After evaluating $f$ at points $\pmb{x}_1,...,\pmb{x}_t$ the predictive posterior $P(f_{t+1} | \pmb{x}_{1:t}, \pmb{y}, \pmb{x}_{t+1}) \sim \mathcal{N}\big(\mu_t(\pmb{x}_{t+1}), cov_t(\pmb{x}_{t+1})\big)$ can be computed in closed form with mean and covariance:
\begin{equation*}
\mu_t(\pmb{x}_{t+1}) = \pmb{k}^T [\pmb{K} + \sigma^2_{noise} \pmb{I}]^{-1} \pmb{y} \quad \quad \quad
cov_t(\pmb{x}_{t+1}) = k(\pmb{x}_{t+1}, \pmb{x}_{t+1}) - \pmb{k}^T [\pmb{K} + \sigma^2_{noise} \pmb{I}]^{-1} \pmb{k},
\end{equation*}
where $\pmb{k} \in \R^t$, with $\pmb{k}_i=k(\pmb{x}_{t+1}, \pmb{x}_i)$; $\pmb{K} \in \R^{t \times t}$ with $\pmb{K}_{ij} = k(\pmb{x}_i, \pmb{x}_j)$; $\pmb{I}$ is an identity $\in \R^{t \times t}$, and $\pmb{y}$ is a vector of values obtained after evaluating $f(\pmb{x}_1), ..., f(\pmb{x}_t)$, assuming Gaussian noise with variance $\sigma^2_{noise}$: $y_i = f(\pmb{x}_i) + \epsilon_{\mathcal{N}(0, \sigma^2_{noise})}$. More details can be found in~\citep{GPsMLBook}.

%===============================================================================

\section{Problem Formulation}
\label{sec:probform}
\label{sec:prob_formulation}
In this work we aim to automatically optimize parameters of controllers for bipedal locomotion with respect to some commonly used cost functions. We assume that for a $d$-dimensional controller there is a bounded region of interest (a hypercube) defined by low/high limits on the values of controller parameters: $\pmb{x} \in [\pmb{x}_{low}, \pmb{x}_{high}] \subset \mathbb{R}^n $. Some parts of this region contain points corresponding to parameter sets of the controller that yield the desired walking behavior. Such regions might comprise a large part of the space with numerous local optima, or might comprise only a small part of the space (e.g. less than 1\%). In other words: we do not impose any overly restrictive assumptions on the space of controller parameters, local/global optima, or structure and properties of the cost functions of interest.

The first setting we consider is the case of optimizing 5-dimensional parameters for Raibert locomotion controller of the ATRIAS robot similar to \citep{martin2017experimental}, \citep{hubicki2016walking} and \citep{martin2015robust}. This controller has a Raibert-like foot placement policy~\citep{raibert1986legged}. It uses a linear feedback law operating on horizontal speed and displacement of the center of mass (CoM) to determine a desired foot touch down point: 
\vspace{-2px}
\begin{equation*}
x_p = k(v - v_{tgt})+C\cdot d + 0.5 \cdot v \cdot T   
\vspace{-2px}
\end{equation*}
Here, $x_p$ is the desired location for the end of swing; $v$ is the current speed of the CoM; $k$ is a feedback term that regulates $v$ towards the target speed $v_{tgt}$; $C$ is a constant and $d$ is the measured distance between the stance leg and the CoM; $T$ is the step time. The term $0.5 \cdot v \cdot T$ is a feedforward term, similar to ~\citep{raibert1986legged}. The swing foot trajectory is defined as a 5th order spline ending at $x_p$.

In stance, we regulate both the torso pitch and CoM height to maintain constant desired values:
\vspace{-2px}
\begin{equation*}
F_x = K_{pt} (\theta_{des} - \theta) + K_{dt}(\dot\theta_{des} - \dot\theta)
\quad \quad F_z = K_{pz} (z_{des} - z) + K_{dz}(\dot z_{des} - \dot z)
\vspace{-2px}
\end{equation*}
These desired forces are sent to an inverse dynamics solver to return the corresponding joint torques that produce these desired ground reaction forces. 
Our 5-dimensional controller consists of $[k,\!C,\!T,\!K_{pt},\!K_{dt}]$.$\!\ $Other parameters can be included, but the performance is not sensitive to them. This controller does not specify a target CoM trajectory. Instead it tries to maintain a constant height and torso angle in stance. The foot-placement strategy determines the resulting motion and speed.

To demonstrate applicability to a challenging setting with a higher-dimensional controller we also experiment with a Neuromuscular model for control~\citep{song2015neural}. Since it has not yet been fully adapted to work on ATRIAS in hardware, for this setting we evaluate our work on a 7-link planar model~\citep{nitishURL}. To facilitate comparison of our results with prior work in~\citep{rai2016sample}, we optimize over a 16-dimensional subspace of controller parameters. Description of the Neuromuscular controller and detailed information about the 16 parameters that are optimized can be found in Section III of~\cite{rai2016sample}. We collect training data from simulations on flat ground. We conduct the evaluation of our approaches on perturbed models to create a simulated mismatch between simulation and hardware. We generate a set of model disturbances for each link of the robot, perturbing the mass, inertia and center of mass location up to $\pm 15\%$ of the original value. In addition, instead of walking on flat ground, we use a set of randomly generated rough ground profiles with step height of up to $\pm 6\ cm$.

%===============================================================================

\section{Proposed Approach}
\label{sec:approach}

The aim of our approach is to automatically learn an informed kernel for optimizing bipedal locomotion controllers with Bayesian Optimization.
An uninformed kernel, like Squared Exponential, operates with vectors that represent controller parameters directly. In contrast, we learn a re-parameterization that incorporates information from simulation. We run short simulations for a range of parameter sets and record the resulting costs from the same cost function as that used in Bayesian Optimization. Costs obtained during short simulations serve as approximate indicators of the quality of the controller parameters. Our idea is to train a neural network to reconstruct the cost landscape of short simulations while focusing on the more promising parts of the space. Section~\ref{sec:approach_asym} describes how this approach yields an informed kernel that helps focus the search on the well-performing regions. We also develop a cost-agnostic approach of reconstructing trajectory summaries instead of cost landscape from short simulations. This is described in Section~\ref{sec:approach_traj}.

\subsection{Regression with Implicitly Asymmetric Loss}
\label{sec:approach_asym}

We consider a cost function focused on matching the desired walking speed and heavily penalizing falls:
\vspace{-11px}
\begin{equation}
cost_{atrias} = 		
    \begin{cases}
		100 - x_{fall} , \text{\small{if fall}} \\
		10 \cdot || \pmb{v}_{tgt} - \pmb{v}_{actual} ||^2, \text{\small{if walk}}\\
	\end{cases}
\label{eq:cost_atrias}
\end{equation}
where $x_{fall}$ is the distance travelled before falling, $\pmb{v}_{tgt}$ is the target velocity and $\pmb{v}_{actual}$ is the vector containing actual velocities of the robot. This kind of cost function is of interest because it helps easily distinguish points that walk from points that fall. Similar costs have been considered in prior work when optimizing locomotion controllers~\citep{song2015neural, rai2016sample}.

\begin{wrapfigure}{r}{0.35\textwidth}
\vspace{-14px}
\caption{\small{2D slice of cost landscape.}}
\vspace{-7px}
\includegraphics[width=0.35\textwidth]{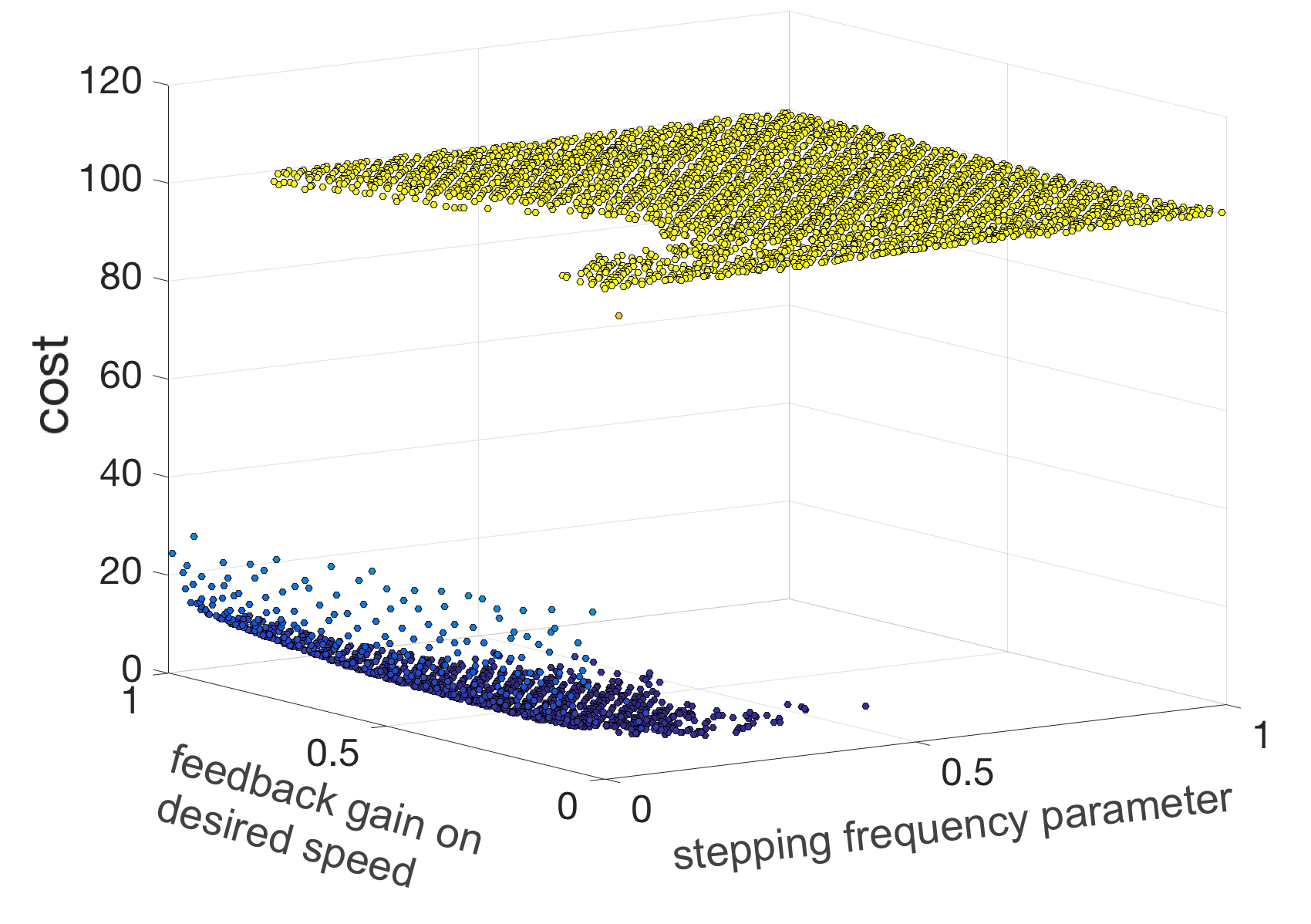}
\label{fig:atrias_cost_landscape}
\vspace{-23px}
\end{wrapfigure}

Figure~\ref{fig:atrias_cost_landscape} shows a scatter plot of applying cost from Equation~\ref{eq:cost_atrias} to simulations of Raibert controller for the ATRIAS robot as introduced in Section~\ref{sec:probform}. For visualization we restrict attention to a 2-dimensional subspace of the parameter space. We pick a well-performing set of parameters (in 5D), then vary the first two dimensions to obtain a 2D subspace. 
The challenge comes from the fact that the boundary between the well-performing (blue) and poorly performing (yellow) parameters is discontinuous. This is a typical landscape for bipedal systems, where a controller that makes the robot fall is much worse than one than walks, and the boundary is extremely sharp. 
Fitting such cost function with regression could be difficult. Learning to reconstruct the boundary exactly using the training set might result in overfitting and poor performance on the test set. Applying regularization is likely to yield high loss and uncertainty about points close to the boundary. This is particularly problematic if poorly performing points lie close to the most promising regions of the parameter space, which is the case in our setting. 

\begin{wrapfigure}{r}{0.26\textwidth}
\centering
\vspace{-10px}
\caption{\small{Cost transform.}}
\vspace{-7px}
\includegraphics[width=0.26\textwidth]{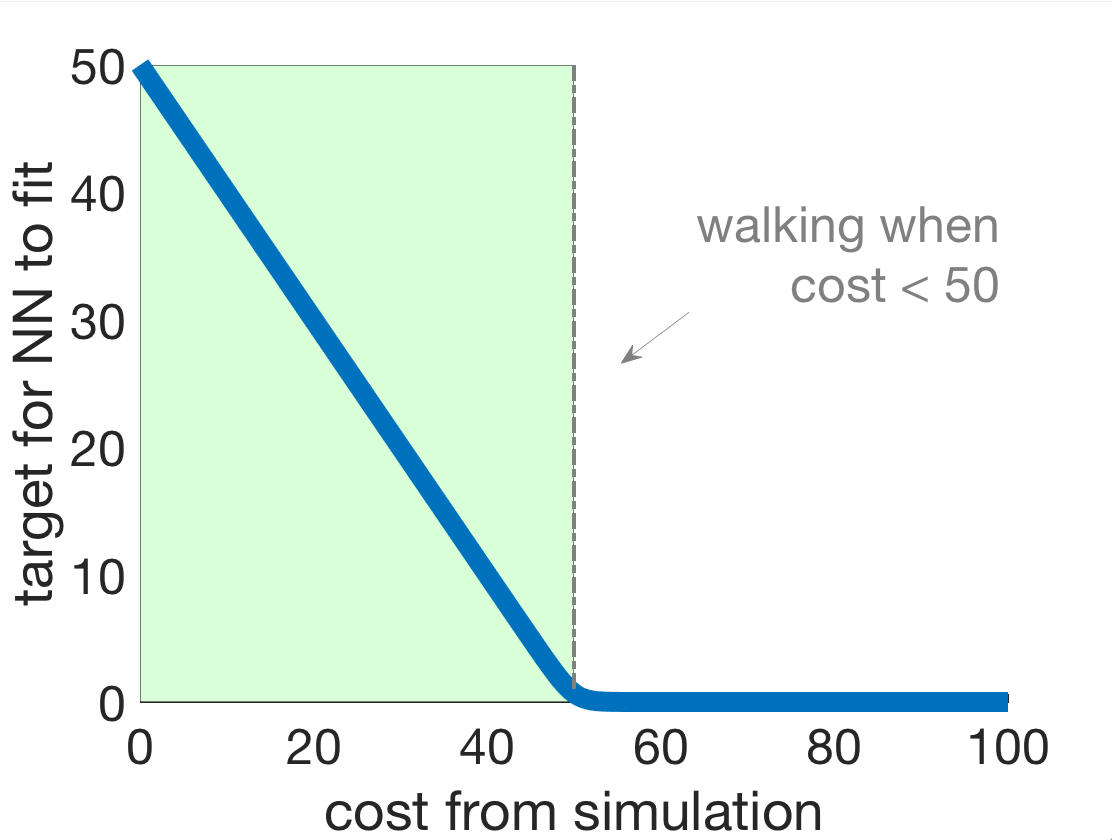}
\label{fig:atrias_cost_transform}
\vspace{-5px}
\end{wrapfigure}

We propose to use a transformation of the cost as the target for the supervised learning. Our approach is to train a deep neural network to reconstruct a reflected shifted softplus function of the cost: 
\begin{equation}
\textit{score}_{\textit{NN}} = \zeta \big( c_{walk} - f_{sim}(\pmb{x}) \big)
\label{eq:score_asymnn}
\end{equation}
Here $\zeta$ is a softplus function: $\zeta(a) = ln\big( 1 + e^a \big)$, $c_{walk}$ is the average cost for the parameter sets that walk during short simulations, $f_{sim}(\pmb{x})$ is the cost computed by the simulator for vector $\pmb{x}$ of controller parameter values. Using this transformation yields a ``score" function such that parameter sets which produce poor results in simulation are mapped to values close to zero. With this, the differences in scores of the poorly performing parameter sets become small or zero. In contrast, the differences in scores of the parameter sets yielding potentially promising results remain proportional to the difference in the corresponding costs. Figure~\ref{fig:atrias_cost_transform} gives a visualization of this transformation.

Cost transformation in Equation~\ref{eq:score_asymnn} serves to essentially create an asymmetric loss for neural network training. This loss is minimized when the promising (low-cost, high-score) points are reconstructed correctly. For the poorly performing (high-cost, low-score) points, it only matters that the output of the neural network is close to zero. Such asymmetric loss can be interpreted as implementing a hybrid of regression and ``soft'' classification. The regression aspect aims to fit the promising points which correspond to walking behaviors. The ``soft'' classification aspect gives an increase in the loss only if a poorly performing point is ``mis-classified'' as well-performing.

When training the neural network we apply L1 loss instead of the usual L2 loss. With this, errors in reconstructing points on the boundary contribute only linearly to the overall loss. This helps achieve a better fit of the stable parts of the parameter space, instead of focusing on the boundary.

We utilize the reconstructed transformed costs to define $\textit{asymNN}$ kernel for Bayesian Optimization:
\vspace{-10px}
\begin{equation}
k_{\textit{asymNN}}(\pmb{x}_i, \pmb{x}_j) = \sigma_k^2 \exp\big(-\tfrac{1}{2 \ell^2} | \textit{score}_{\textit{NN}}(\pmb{x}_i) - \textit{score}_{\textit{NN}}(\pmb{x}_j) |^{2} \big)
\end{equation}
with hyperparameters $\sigma_k^2, \ell$ as described in Section~\ref{subsec:bo_background}.
The proposed approach is able to clearly separate the unpromising part of the parameter space. Under the resulting distance metric poorly performing sets of parameters are close together and far from well-performing ones. 

\subsection{Reconstructing Cost-agnostic Trajectory Summaries}
\label{sec:approach_traj}

Utilizing costs obtained from short simulations provides a way to build an informed kernel without specifying any additional domain knowledge. 
If simulations are computationally expensive it is desirable to minimize the need to repeat data collection. Often, the cost function needs to be modified to accommodate different objectives, hence, there is a need for a cost-agnostic approach. For such cases we propose to train a neural network to reconstruct summaries of trajectories that are cost-agnostic, then utilize these trajectory summaries for constructing kernel distance metric.

We summarize trajectories by recording fairly generic aspects of locomotion: walking time (time before falling), energy used during walking, position of the torso, angle of the torso, coordinates of the center of mass at the end of the short simulation runs. These summaries of simulated trajectories are collected for a range of controller parameters and comprise the training set for the neural network to fit (input: $\pmb{x}$ -- a set of control parameters; output: $\pmb{traj_x}$ -- the corresponding trajectory summary obtained from simulation). The outputs of the (trained) neural network
offer the reconstructed/approximate trajectory summaries:
$f_{\textit{NN}}(\pmb{x}) = \pmb{\widehat{traj}_x}$, where $\pmb{x}$ is the input controller parameters, and $\pmb{\widehat{traj}_x}$ is the corresponding reconstructed trajectory summary.
These are then used to construct an informed cost-agnostic kernel for Bayesian Optimization:
\begin{equation}
k_{\textit{trajNN}}(\pmb{x}_i, \pmb{x}_j) = \sigma_k^2 \exp\big(\!-\!\tfrac{1}{2} \pmb{t}_{ij}^T \diag( \pmb{\ell})^{\!-\!2} \pmb{t}_{ij}\big), \quad \quad \pmb{t}_{ij} = f_{\textit{NN}}(\pmb{x}_i) - f_{\textit{NN}}(\pmb{x}_j) 
\end{equation}
The general concept of utilizing trajectory data to improve sample efficiency of BO has been proposed before, for example in~\citep{wilson2014using}. However, prior work assumed obtaining trajectory data is possible every time kernel values $k(\pmb{x}_i, \pmb{x}_j)$ need to be evaluated. This is not the case in our setting. Trajectory summaries are initially obtained via costly high-fidelity simulations, and it would be infeasible to compute trajectory information via simulation during BO. Hence, our approach is to train a neural network to learn reconstructing trajectory summaries first. Running a forward pass of the neural network is a relatively inexpensive operation, hence reconstructed/approximate trajectory summaries can be quickly obtained during BO whenever $k_{\textit{trajNN}}(\pmb{x}_i, \pmb{x}_j)$ needs to be computed. 

When defining trajectory summaries we did not focus on carefully selecting what aspects to include/exclude. Our goal was an approach that could be quickly adapted to other domains. When applying this approach to a new domain, the strategy could be simply to include trajectory information used to compute cost functions that are of interest/relevance in the domain. For example, for a manipulator, the coordinates of  end-effector(s) could be recorded at relevant points. In principle, our approach also could utilize domain- and task-specific `descriptors', like those proposed in~\citep{rai2016sample, cully2015robots}.

%===============================================================================

\section{Experimental Results}
\label{sec:experiments}

In this section we describe our experiments with cost-based and trajectory-based kernels. We first consider the setting of optimizing a 5-dimensional controller for the ATRIAS robot. We show that the cost-based kernel is able to improve sample efficiency over standard Bayesian Optimization. We present hardware experiments to demonstrate that our kernel allows obtaining a set of parameters close to optimal on the second trial. We then discuss simulation experiments with a 16 dimensional controller that utilizes a Neuromuscular model~\citep{song2015neural}. These experiments show that our trajectory-based kernel is able to significantly outperform standard Bayesian Optimization for a higher-dimensional controller even when a sharply discontinuous cost is used during optimization.

\subsection{Experiments with Raibert controller on the ATRIAS robot}
\label{experiments_atrias}

For our experiments on the ATRIAS robot we used a high-fidelity ATRIAS simulator~\citep{martin2015robust} to generate the kernel. We did an initial analysis of the performance of our approach in simulation, followed by hardware experiments. 
As described in Section~\ref{sec:approach_asym}, we trained a neural network to reconstruct cost information obtained from short simulations. We created a sobol grid on the input parameter space with 20K points and ran short 3.5 second simulations on each of the corresponding 20K parameter sets to compute the costs. We then trained a fully connected network with 4 hidden layers (128, 64, 16, 4 units) to reconstruct $\textit{score}_{\textit{NN}}$ (the transformation of the cost described in Section~\ref{sec:approach_asym}).

\begin{wrapfigure}{r}{0.39\textwidth}
\vspace{-12px}
\caption{\small{Initial tests of Bayesian Optimization for 5-dimensional controller in simulation. The plot shows mean of best cost so far over 30 runs for each kernel, error bars are 95\% confidence intervals.}}
\includegraphics[width=0.39\textwidth]{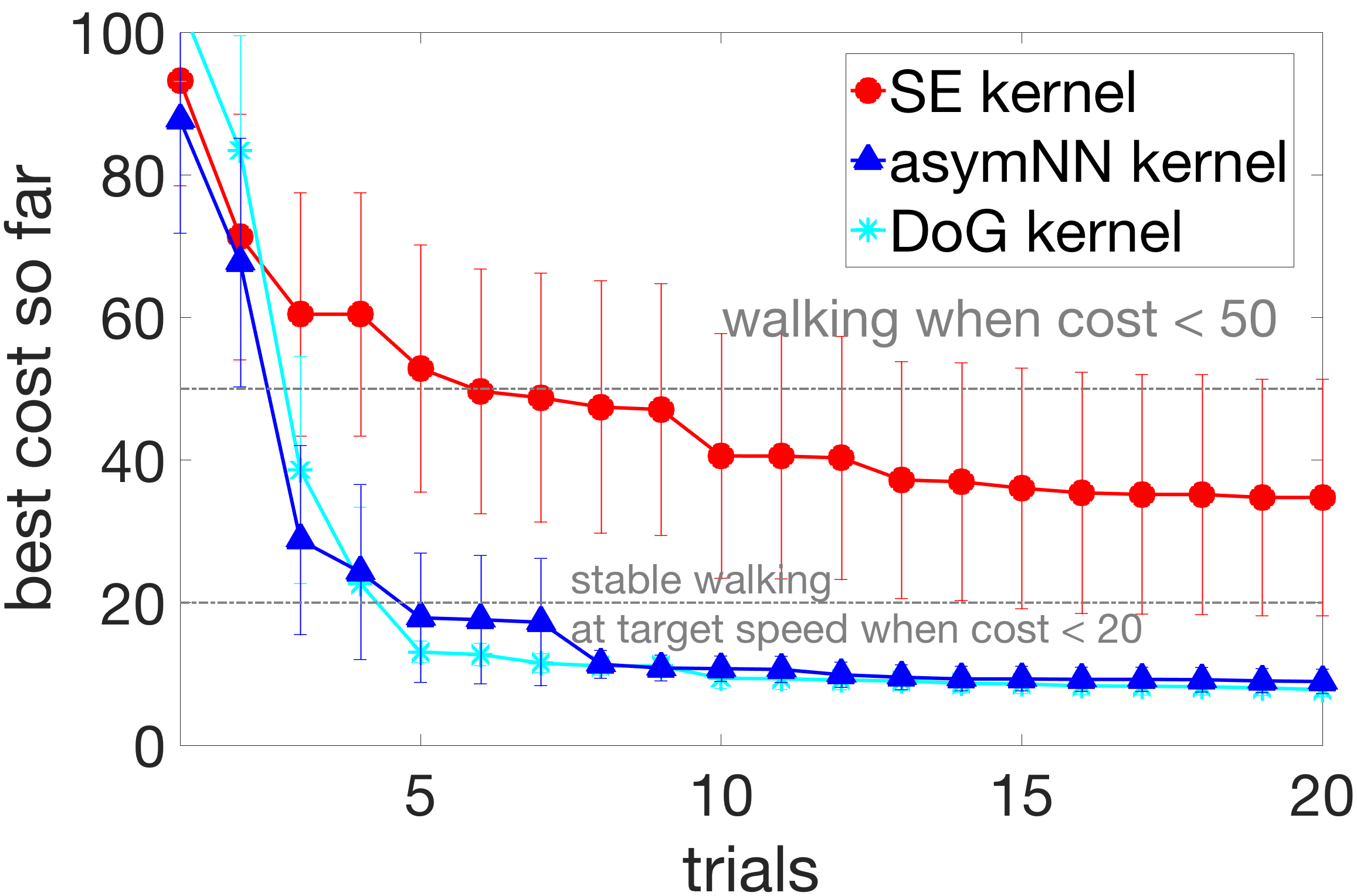}
\label{fig:bo_runs_atrias_sim}
\vspace{-18px}
\end{wrapfigure}

In Figure~\ref{fig:bo_runs_atrias_sim} we first compare the performance of BO that used our neural network-based kernel (\textit{asymNN}) versus using a standard Squared Exponential kernel (\textit{SE}) in simulation. For these experiments we used the cost from Equation~\ref{eq:cost_atrias}, Section~\ref{sec:approach_asym} with target velocity of $1m/s$. Simulations with cost less than 50 yielded walking behavior, those with cost less than 20 yielded stable walking close to the target speed. BO with \textit{asymNN} kernel reliably found stable walking points after only 8 trials. In contrast, BO with \textit{SE} kernel did not find stable walking solutions in the first 20 trials reliably. We also compare with a recently proposed Determinants of Gait (\textit{DoG}) kernel~\citep{rai2016sample}. \textit{DoG} utilized more specific domain knowledge to construct an informed kernel for BO of locomotion controllers for a fixed set of points. \textit{asymNN} was able to closely match the performance of \textit{DoG} in this setting after 8 trials.

After experiments in simulation suggested that \textit{asymNN} kernel can yield a significant improvement in sample efficiency of BO, we conducted a set of experiments on the ATRIAS robot. We completed 6 sets of runs of BO: 3 using \textit{asymNN} kernel and 3 using a standard \mbox{\textit{SE} kernel} with 10 trials each, leading to a \mbox{total of 60 hardware experiments}.

Since ATRIAS walks around a rather short boom in 2D, walking at high speeds needs a lot of torque from the robot motors. This means higher lateral forces between the robot and the boom, which

\begin{wrapfigure}{r}{0.5\textwidth}
\centering
\caption{\small{ATRIAS during BO with \textit{asymNN} kernel.}}
\vspace{-5px}
\includegraphics[width=0.49\textwidth]{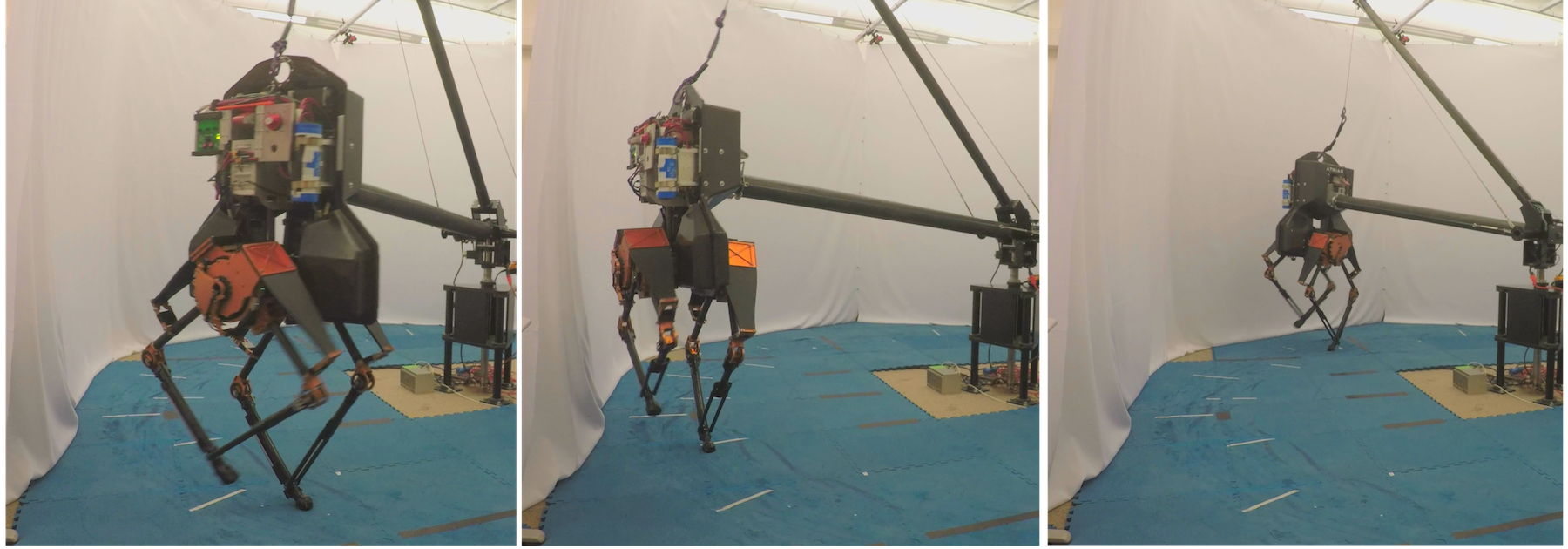}
\label{fig:bo_runs_atrias_hw_slides}
\vspace{-10px}
\end{wrapfigure}
\mbox{do not} affect our direction of motion but can lead to a lot of internal forces, eventually breaking the robot.
So, in our first attempt, we tried to start with lower speeds of $0.4m/s$ so that we could do hardware experiments and analyze the validity of our approach on hardware without breaking the robot too often. In this setting with low target speed, stable walking points comprised $\approx\!\!\frac{1}{6}$ of the parameter space. We anticipated it would be challenging to improve over BO with \textit{SE} kernel, since it was able to find stable walking solutions after only 3-4 trials.

\begin{figure}[t]
\centering
\caption{\small{Hardware experiments on the ATRIAS robot.}}
\vspace{-5px}
\begin{subfigure}[t]{0.46\textwidth}
\centering
\caption{\small{Best cost so far during BO (mean over 3 runs, \\ shaded region indicates $\pm$ one standard deviation).}}
\vspace{-5px}
\includegraphics[width=1.0\textwidth]{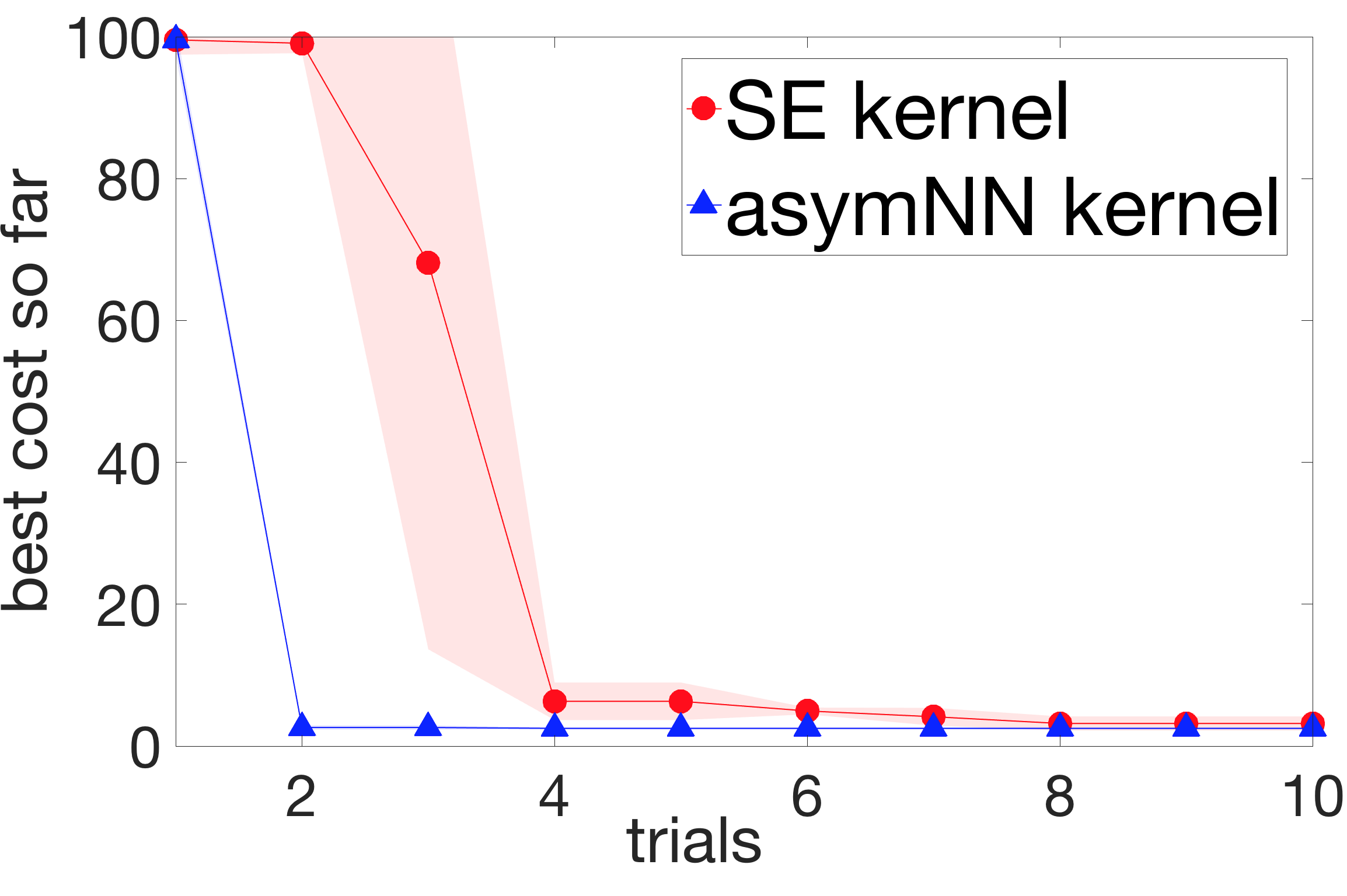}
\label{fig:bo_runs_plot_atrias_hw}
\end{subfigure}
\quad
\begin{subfigure}[t]{0.46\textwidth}
\centering
\caption{\small{Number of ``walking'' points sampled \\ (out of 10 trials in each run).}}
\vspace{-3px}
\includegraphics[width=0.85\textwidth]{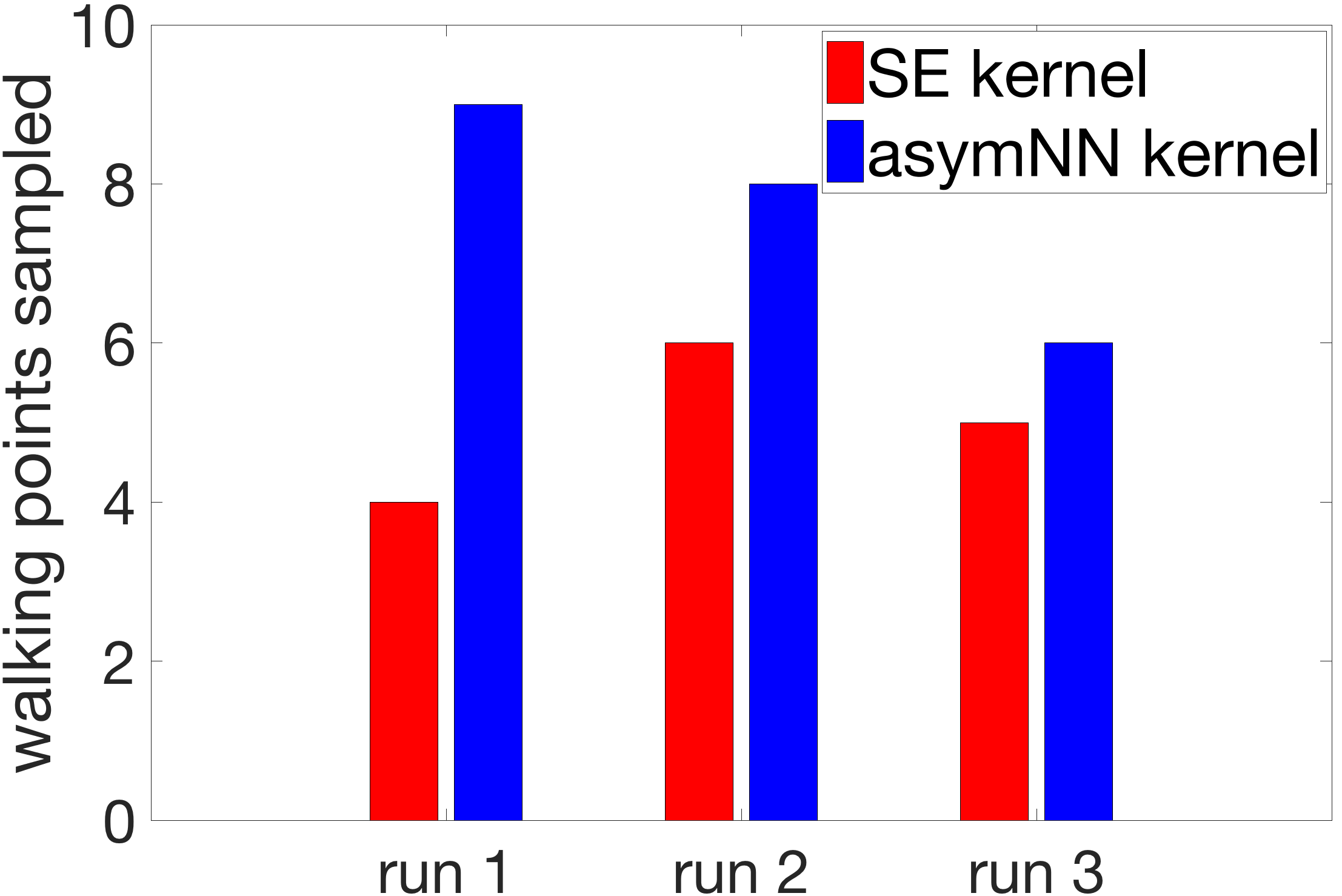}
\label{fig:bo_runs_hist_atrias_hw}
\end{subfigure}
\vspace{-35px}
\end{figure}

Figure~\ref{fig:bo_runs_plot_atrias_hw} shows the performance of BO with SE versus \textit{asymNN} kernel. 
SE obtains a stable walking solution on the 3rd trial in one run, and on the 4th trial in the two other runs. \textit{asymNN} kernel is able to find the best-performing set of parameters on the second trial in each of the 3 runs. This confirms that using \textit{asymNN} kernel offers an improvement over using \textit{SE} kernel in this setting. We suggest that \textit{asymNN} reliably selects an excellent point on the 2nd trial  because such points lie far from  poorly performing subspace of parameters (under the distance metric constructed with \textit{asymNN}). 
\textit{asymNN} kernel also helped sampling more walking points overall (Figure \ref{fig:bo_runs_hist_atrias_hw}). This is desirable as stable points are less likely to break the robot.

While in the above hardware setup most methods are likely to sample walking points within 10 trials, we believe our experimentation is an important step towards optimizing locomotion policies for complex humanoid robots. BO studies in the past also used real robot hardware (e.g. \citep{calandra2016AMAI, cully2015robots, tesch}). However, \citep{tesch, lizotte2007automatic, cully2015robots} used robots which are statically stable for significant parts of their gait, making discontinuities in the cost function landscape less likely and in turn making the optimization easier. In contrast, ATRIAS is a complex bipedal system which is likely to fall with unstable controllers due to point feet. \citep{calandra2016AMAI} use a walking robot similar to ours. However, their controller parametrization is very different, and not widely used, unlike our inverse dynamics and force-based controller which is more modern and state-of-the-art \citep{kuindersma2016optimization}, \citep{herzog2016momentum}, \citep{feng2015optimization}. 
While with our hardware setting it might be hard to show improvement over simpler approaches at low constant target speeds, we believe the setting is adequate, because our testbed is fairly complex and our problem formulation is widely applicable. 
In Appendix A we describe initial results for variable target speed experiments, with SE kernel not finding walking solutions reliably even after 20 trials and asymNN succeeding after the first 10 trials.

\subsection{Experiments with the Neuromuscular Model}
\label{experiments_nm}

16-dimensional controller of the Neuromuscular model (described in Section~\ref{sec:prob_formulation}) yielded a challenging optimization setting: walking points comprised less than 2\% of the parameter space in simulation. Here we describe our experiments with cost-agnostic approach for constructing an informed kernel introduced in Section~\ref{sec:approach_traj}. We created a grid of 100K points in the input parameter space  and ran short 5 second simulations on each of the corresponding 100K parameter sets to collect the trajectory summaries. We then trained a fully connected network with 3 hidden layers (512, 128, 32 units) with L1 loss to reconstruct 8-dimensional trajectory summaries (as described in Section~\ref{sec:approach_traj}). All experiments were done on perturbed models, as described in Section~\ref{sec:probform}.

\begin{figure}[t]
\centering
\caption{\small{Bayesian Optimization for the Neuromuscular model controller in simulation. 
\textit{trajNN} and \textit{DoG} kernels were constructed with undisturbed model on flat ground.
BO is run with mass/inertia disturbances on different rough ground profiles to simulate mismatch.
Plots show means over 50 runs, 95\% confidence intervals.}}
\vspace{-3px}
\begin{subfigure}[t]{0.48\textwidth}
\centering
\caption{\small{Using smooth cost from Equation~\ref{eq:cost_smooth}.}}
\vspace{-5px}
\includegraphics[width=0.9\textwidth]{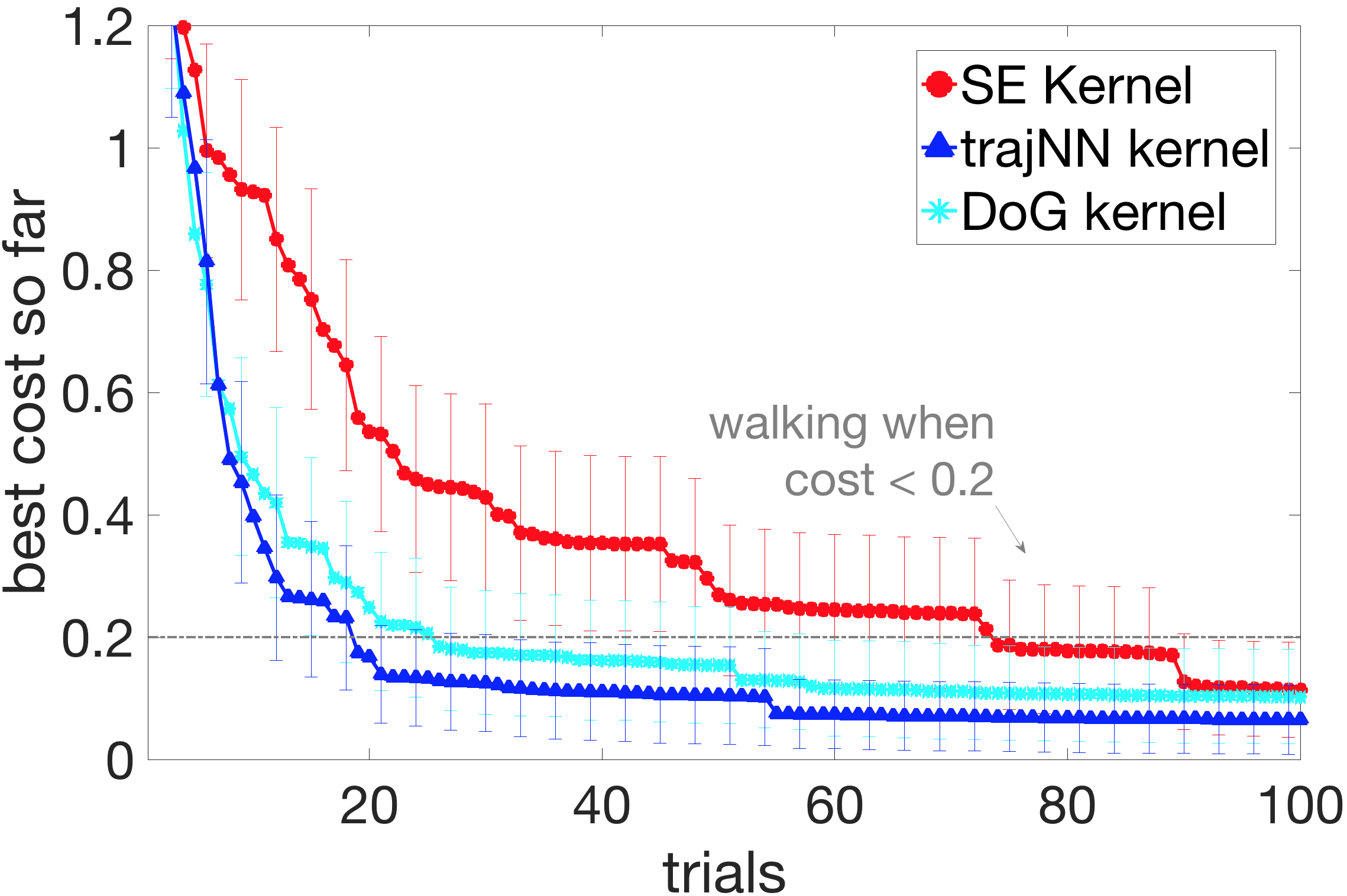}
\label{fig:smooth_cost_bo_runs}
\end{subfigure}
\quad
\begin{subfigure}[t]{0.48\textwidth}
\centering
\caption{\small{Using non-smooth cost from Equation~\ref{eq:cost_nonsmooth}.}}
\vspace{-5px}
\includegraphics[width=0.9\textwidth]{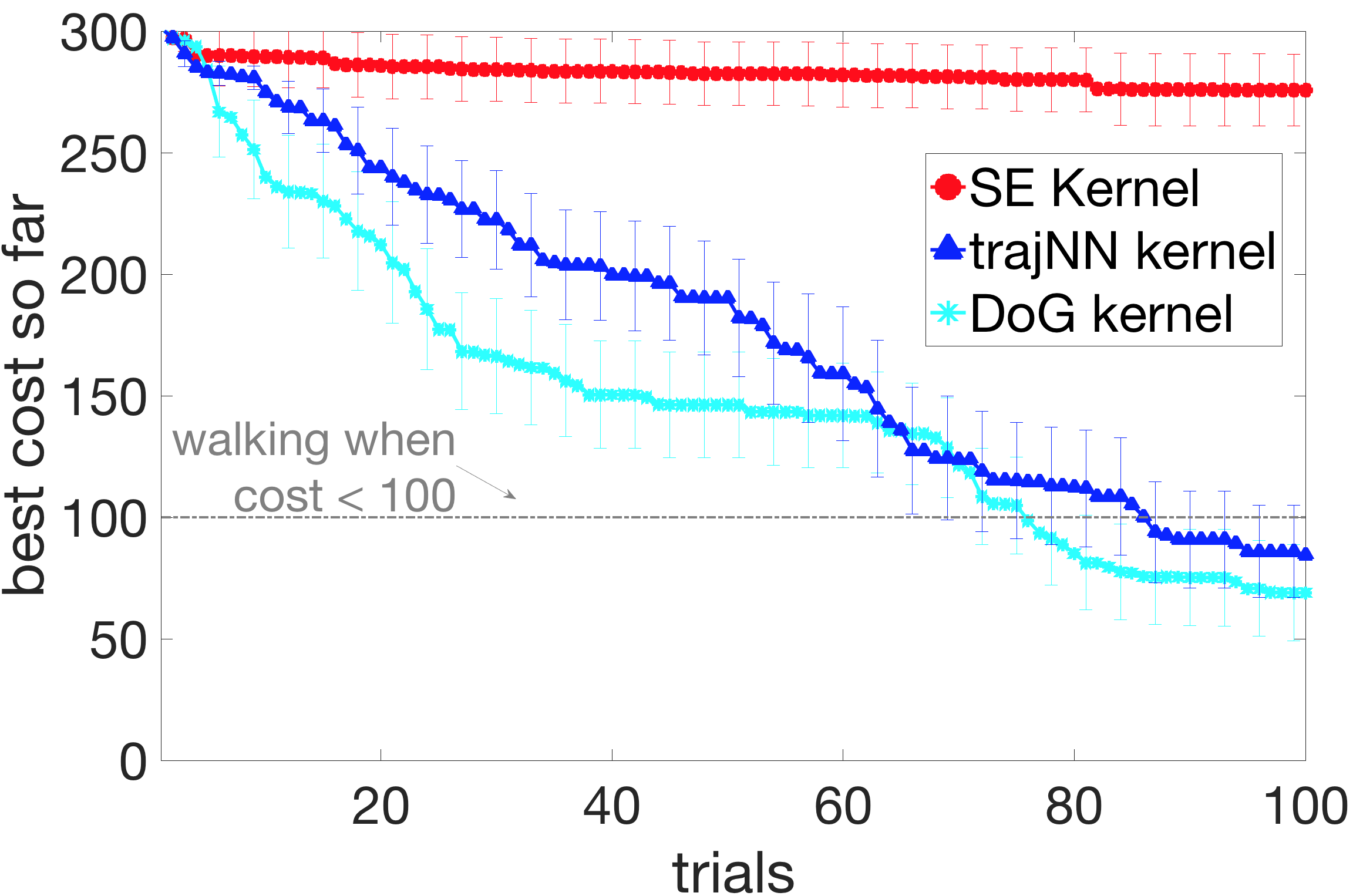}
\label{fig:nonsmooth_cost_bo_runs}
\end{subfigure}
\label{fig:bo_runs_nm}
\vspace{-10px}
\end{figure}

Figure~\ref{fig:bo_runs_nm} compares using \textit{trajNN} versus \textit{SE} kernel for BO with two different costs from prior literature. The first cost promotes walking further and longer before falling, while penalizing deviations from the target speed~\citep{rai2016sample}:
\begin{equation}
cost_{smooth} = 1/(1+t) + 0.3/(1+d) + 0.01(s-s_{tgt}),
\label{eq:cost_smooth}
\end{equation}
where $t$ is seconds walked, $d$ is the final hip position, $s$ is mean speed and $s_{tgt}$ is the desired walking speed ($1.3m/s$ in our case). 
The second cost function is a simplified version of the cost used in~\citep{song2015neural}. It penalizes falls explicitly, and encourages walking at desired speed and with lower cost of transport:
\vspace{-3px}
\begin{equation}
cost_{non\text{-}smooth} = 		
    \begin{cases}
		300 - x_{fall} , \text{\small{if fall}} \\
		100 ||v_{avg} - v_{tgt}|| + c_{tr}, \text{\small{if walk}}\\
	\end{cases}
\label{eq:cost_nonsmooth}
\end{equation}
where $x_{fall}$ is the distance covered before falling, $v_{avg}$ is the average speed of walking, $v_{tgt}$ is the target velocity, and $c_{tr}$ captures the cost of transport.

Figure~\ref{fig:smooth_cost_bo_runs} shows that \textit{trajNN} offers a significant improvement in sample efficiency when using $cost_{smooth}$. Points with cost less than $0.2$ correspond to robust walking behavior. With \textit{trajNN}, more than $90\%$ of runs obtain walking solutions after only 25 trials. In contrast, using \textit{SE} requires more than 90 trials for such success rate. The performance of \textit{trajNN} matches that of a \textit{DoG} kernel from prior work~\citep{rai2016sample}. This is notable, since \textit{trajNN} is learned automatically, whereas \textit{DoG} kernel is constructed using domain expertise.
Figure~\ref{fig:nonsmooth_cost_bo_runs} shows that \textit{trajNN} also provides a significant improvement when using the second cost. Points with cost less than $100$ correspond to walking. With \textit{trajNN}, 70\% of the runs find walking solutions after 100 trials. In contrast, optimizing non-smooth cost is very challenging for BO with \textit{SE} kernel: a walking solution is found only in 1 out of 50 runs after 100 trials.
The difference in performance on the two costs is due to the nature of the costs. For example, if a point walks some distance $d$, Equation \ref{eq:cost_smooth} includes a term $\tfrac{1}{d}$ and Equation \ref{eq:cost_nonsmooth} includes $-d$. A sharper fall in the first cost causes BO to exploit around points that walk some distance. It then quickly finds points that walk forever, while BO with the second cost continues to explore.

%===============================================================================

\section{Conclusion and Future Work}
\label{sec:conclusion}
In this work we proposed learning informed kernels for Bayesian Optimization of locomotion controllers without relying heavily on domain experts. We optimized a 5-dimensional controller on the ATRIAS robot and showed that our cost-based kernel offered an improvement over using an uninformed kernel. We also proposed a cost-agnostic alternative. Experiments with a 16-dimensional Neuromuscular controller in simulation showed a significant improvement with different costs. 

In future work it would be interesting to further analyze various approaches that enhance sample efficiency of BO. Approaches that embed simulation-based information into the kernel (like those we proposed) can enhance sample efficiency dramatically by focusing BO on regions that look promising in simulation. Approaches that use simulation-based samples in BO posterior mean directly (e.g. \cite{marco2017virtual}) could be more robust to simulation-based inaccuracies after collecting a larger amount of data from hardware experiments. However, they can only incorporate cost-based information (e.g. no way to add trajectory information directly to the posterior mean). Perhaps there is an effective way to combine the two directions.
Another promising line for future work is learning flexible models of simulation-vs-hardware mismatch. Such models could help decrease the influence of distortion from incorrect simulations and could help enhance both `kernel-based' and `mean-based' methods.

%===============================================================================

% The maximum paper length is 8 pages excluding references and acknowledgements, and 9 pages including references and acknowledgements

\clearpage
% The acknowledgments are automatically included only in the final version of the paper.
\acknowledgments{\thanks{This research was supported in part by National Science Foundation grant IIS-1563807, the Max-Planck-Society, \& the Knut and Alice Wallenberg Foundation. Any opinions, findings, and conclusions or recommendations expressed in this material are those of the author(s) and do not necessarily reflect the views of the funding organizations.}}

%===============================================================================

% no \bibliographystyle is required, since the corl style is automatically used.
\bibliography{boforloconn}  % .bib

%===============================================================================
\appendix
\label{sec:appendix}
\section{Hardware Experiments with Variable Speed Profile}

\begin{wrapfigure}{r}{0.47\textwidth}
\vspace{-10px}
\centering
\caption{\small{Hardware experiments on the ATRIAS robot with variable target speed profile ($0.4m/s$ - $1.0m/s$ - $0.2m/s$). Plot shows best cost so far during BO (mean over 3 runs, shaded region shows $\pm$~one~standard~deviation).}}
\vspace{-5px}
\includegraphics[width=0.45\textwidth]{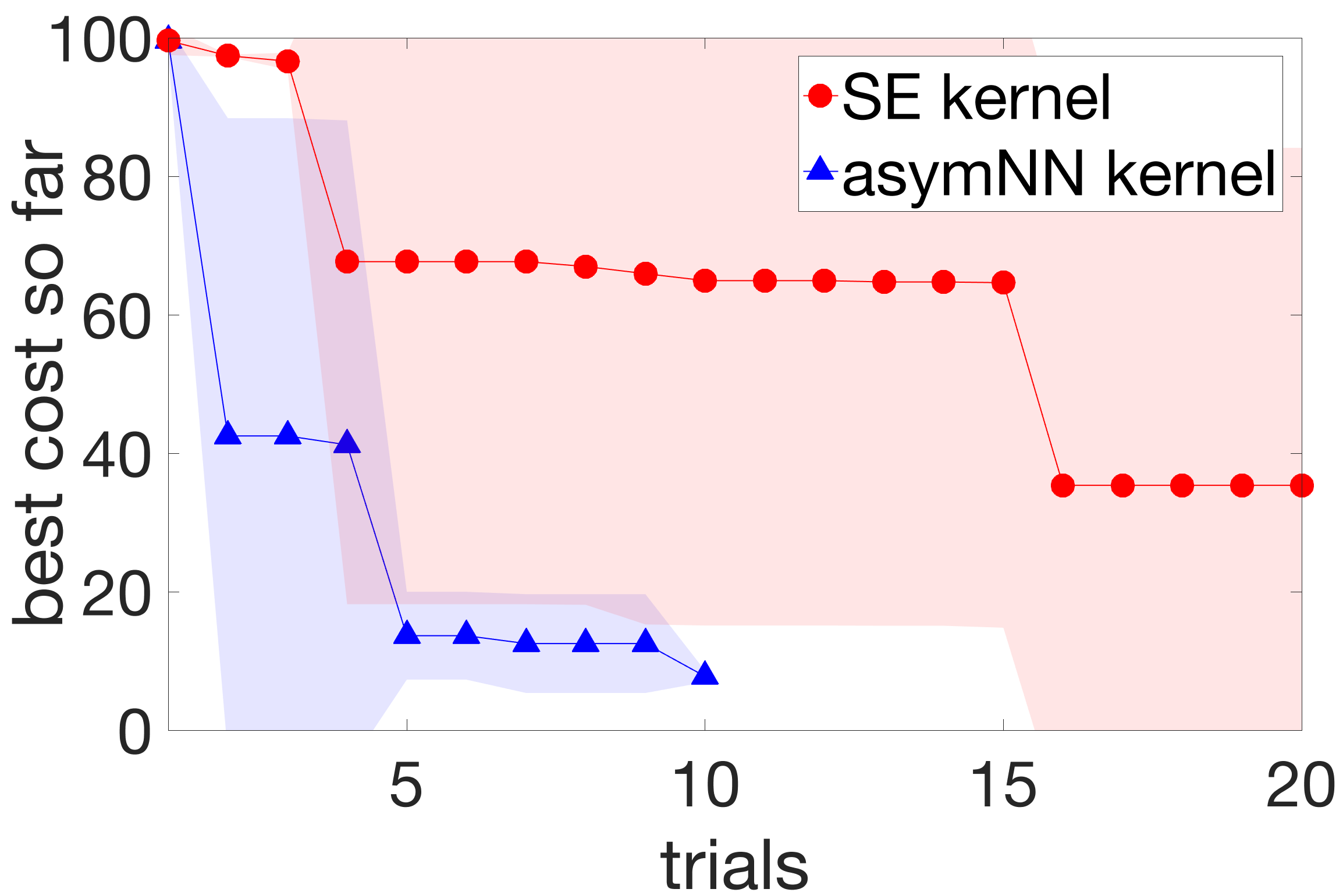}
\label{fig:hw_varspeed_pm_1std}
\vspace{-10px}
\end{wrapfigure}
In this set of experiments we used a variable target speed profile on the hardware setting described in Section~\ref{experiments_atrias}. The target was for ATRIAS to walk at the speed of $0.4m/s$ for 15 steps, then speed up to $1.0m/s$ (15 steps), then slow down to $0.2m/s$ (15 steps). This setting was still safe enough to avoid frequent hardware malfunction, and yet more challenging than walking at a constant speed of $0.4m/s$.
Figure~\ref{fig:hw_varspeed_pm_1std} shows the performance of BO with SE versus \textit{asymNN} kernel. As with experiments in Section~\ref{experiments_atrias}, we completed 6 sets of runs of BO: 3 using \textit{asymNN} kernel and 3 using a standard SE kernel with 10 trials each, leading to a total of 60 hardware experiments. BO with SE kernel did not find walking solutions reliably even after 20 trials. In contrast, \textit{asymNN} succeeded after the first 10 trials in all of the runs.
These new sets of experiments show that the \textit{asymNN} generalizes to more difficult settings where the standard SE kernel fails to find walking points reliably. We intend to continue experimenting with other higher-dimensional controllers in the future to further test the limits of the neural network based kernels described in this work.

One technical issue was that our \textit{asymNN} kernel runs for variable target speed were done after new flooring was installed. While the influence is small, different flooring affects the performance of controllers as some lower level parameters have to be tuned again. For example, the friction coefficients were different between the two floor mats, as well as their stiffness. So to finalize the comparison, BO with SE kernel runs would need to be re-run with the new environment conditions (though we expect very little change in the performance of both algorithms). To judge the overall increase in difficulty for the variable speed setting it would be necessary to complete a set of experiments testing randomly selected controller parameters as well (as we did for results presented in Section~\ref{experiments_atrias}). We intend to continue this line of experiments in the future.

%===============================================================================
\end{document}